\documentclass{article}

\PassOptionsToPackage{square,numbers}{natbib}
\usepackage[preprint]{neurips_2022}

\usepackage[utf8]{inputenc} %
\usepackage[T1]{fontenc}    %
\usepackage{url}            %
\usepackage{booktabs}       %
\usepackage{amsfonts}       %
\usepackage{nicefrac}       %
\usepackage{microtype}      %
\usepackage{xcolor}         %

\newtheorem{example}{Example}
\usepackage[pdftex]{graphicx}
\usepackage{comment}
\usepackage{color, colortbl}
\definecolor{Gray}{gray}{0.9}

\usepackage{floatrow}
\newfloatcommand{capbtabbox}{table}[][\FBwidth]
\usepackage{blindtext}
\usepackage{floatrow}

\title{Explainable Reinforcement Learning via Model Transforms}%

\author{Mira Finkelstein\textsuperscript{\rm 1},  Lucy Liu\textsuperscript{\rm 2}, Nitsan Levy Schlot\textsuperscript{\rm 1}, Yoav Kolumbus\textsuperscript{\rm 1},\\  {\bf David C. Parkes}\textsuperscript{\rm 2,3}, {\bf Jeffrey S. Rosenschein}\textsuperscript{\rm 1} and {\bf Sarah Keren}\textsuperscript{\rm 4}\\
    \textsuperscript{\rm 1} The Hebrew University of Jerusalem, Benin School of Computer Science and Engineering\\
    \textsuperscript{\rm 2} Harvard University, School of Engineering and Applied Sciences\\
    \textsuperscript{\rm 3} DeepMind\\
    \textsuperscript{\rm 4} Technion - Israel Institute of Technology, Taub Faculty of Computer Science
 }

\newcount\Comments  % 0 suppresses notes to selves in text
\definecolor{darkgreen}{rgb}{0.0, 0.5, 0.0}
\newcommand{\kibitz}[2]{\ifnum\Comments=1{\color{#1}{#2}}\fi}

\newcount\CommentsAdd  % 0 suppresses notes to selves in text
\newcommand{\kibitzAdd}[2]{\ifnum\CommentsAdd=1{\color{#1}{#2}}\fi}

\newtheorem{definition}{Definition}

\usepackage[normalem]{ulem}
\usepackage{enumitem}

\newcommand{\states}{S}

\newcommand{\initState}{s_{0}}
\newcommand{\actions}{A}
\newcommand{\rewardFunc}{R}
\newcommand{\transitionFunc}{P}
\newcommand{\discountFactor}{\gamma}
\newcommand{\RLPE}{Reinforcement Learning Policy Explanation}
\newcommand{\RLPEACR}{RLPE}
\newcommand{\RLPEModel}{R}

\newcommand{\MDP}{M}

\newcommand{\agent}{A} 

\newcommand{\transforms}{T} 
\newcommand{\policy}{\pi}
\newcommand{\policies}{\Pi}
\newcommand{\partialPolicy}{\tilde{\pi}}

\newcommand{\actor}{actor}
\newcommand{\observer}{observer}
\newcommand{\explainer}{explainer}
\newcommand{\antpolicy}{anticipated policy}

\newcommand{\satisfies}{\models}

\newcommand{\allMDPs}{\mathcal{M}}
\newcommand{\transform}{T}
\newcommand{\weightingFunc}{w}

\newcommand{\transformSeq}{\vec{\transform}}
\newcommand{\transformsUniversal}{\mathcal{T}}

\newcommand{\stateAbsFunc}{\phi}
\newcommand{\actionAbsFunc}{\psi}

\newcommand{\policyMR}{satisfaction ratio}

\newcommand{\RL}{RL}
\newcommand{\DRL}{DRL}

\newcommand{\pre}{pre}

\newcommand{\MDPdistance}{d}

\newcommand{\statesFunc}{\mathbb{S}}

\newcommand{\jointActions}{\mathcal{A}}

\begin{document}

\maketitle
\begin{abstract}
Understanding emerging behaviors of reinforcement learning (RL) agents may be difficult since such agents are often trained in complex environments using highly complex decision making procedures.
This has given rise to a variety of approaches to {\em explainability} in RL that aim to reconcile discrepancies that may arise between the behavior of an agent and the behavior that is anticipated by an observer. 
Most recent approaches have relied either on domain knowledge that may not always be available, on an analysis of the agent's policy, or on an analysis of specific elements 
of the underlying  environment, typically modeled as a Markov Decision Process (MDP). Our key claim is that even if the underlying model is not fully known (e.g., the transition probabilities have not been accurately learned) or is not maintained by the agent (i.e., when using model-free methods), the model %\dcp{what?}
can nevertheless be exploited to automatically generate explanations. For this purpose, we suggest using formal MDP abstractions and transforms, previously used in the literature  for expediting the search for optimal policies, to automatically produce explanations.
Since such transforms are typically based on a symbolic representation of the environment, they can provide meaningful explanations for gaps between the anticipated and actual agent behavior.  
We formally define the explainability problem,
%\dcp{which problem?},
suggest a class of transforms that can be used for explaining emergent behaviors, and suggest methods that enable efficient search for an explanation. We demonstrate the approach on a set of standard benchmarks.

\end{abstract}
\section{Introduction}
The performance-transparency trade-off
is a major challenge with many artificial intelligence (AI) methods: as the inner workings of an agent's decision making procedure %
increases in complexity, it becomes more powerful, but the agent's
decisions become harder to understand. 
Accordingly, interest in {\em explainable AI} and the development of transparent, interpretable, AI models has increased rapidly in recent years \cite{XAI20}.
This increase in complexity is particularly prevalent in {\em reinforcement learning} (\RL) and {\em deep reinforcement learning} (\DRL), where an agent autonomously learns how to operate in its environment. 
While \RL~has been successfully applied to solve many challenging tasks, including traffic control \cite{traffic10}, robotic motion planning \cite{robo13}, and board games \cite{schrittwieser2020mastering}, it is increasingly challenging to explain the behavior of RL agents, especially when they do not operate as anticipated.
To allow humans to collaborate effectively with RL-based AI systems and increase their usability, it is therefore important to develop automated methods for reasoning about and explaining agent behaviors.

While there has been recent work on explainability of DRL (see \cite{surv20} for a recent survey), most of these methods either rely on domain knowledge, which may not be available, or involve post-processing the policy learned by the agent (e.g., by reasoning about the structure of the underlying neural network \cite{DBLP:journals/corr/abs-1711-00138}). Moreover, most existing methods for explainability do not fully exploit the formal model that is assumed to represent the underlying environment, typically a Markov Decision Process (MDP) \cite{bertsekas1995dynamic}, and 
analyze instead one chosen element of the model (e.g., the reward function \cite{rewDec19}). 

We focus on RL settings in which the model of the underlying environment may be partially known, i.e., the state space and action space are specified, but the transition probabilities and reward function are not fully known.  This is common to many RL settings in which the action and state spaces are typically known but the agent must learn the reward function and transition probabilities, either explicitly as in model-based RL or implicitly as when learning to optimize its behavior in model-free RL. For example, in a robotic setting, the agent may have some representation of the state features (e.g., the location of objects) and of the actions it can perform (e.g., picking up an object), but  not know its reward function or the probabilities of action outcomes.  

Our key claim is that even if the underlying model is not fully known (or not explicitly learned), it can nevertheless be used to automatically produce meaningful explanations for the agent's behavior, i.e., even if the agent is using a model-free method, the partial model can be manipulated using a model-based analysis to produce explanations.
Specifically, we suggest producing explanations by searching for a set of formal abstractions and transforms that when applied to the (possibly incomplete or approximate) %\dcp{approximately learned (?)} 
MDP representation will yield a behavior that is aligned with an observer's expectations. %\dcp{does this mean you need to work with model-based approaches? if so, the earlier ref to model-free can be dropped} 
For this purpose, we exploit the rich body of literature that offers  MDP transforms \cite{hoffmann2001ff,bonet2005mgpt,tow06,DBLP:journals/jair/Helmert06,geffner2013concise,lif18} that manipulate different elements of the model by, for example, ignoring the stochastic nature of the environment, ignoring some of the effects of actions, and removing or adding 
constraints. 
%\dcp{considering, or removing/ignoring? also, drop `irrelevant`?} 
 While these methods have so far been used to expedite planning and learning, we use them to automatically produce explanations. That is, while for planning the benefit of using such transforms is in increasing solution efficiency, we use them to isolate features of the environment model that cause an agent to deviate from a behavior that is anticipated by an observer.    

Formally, we consider an explainability setting, which we term \emph{\RLPE}
(\RLPEACR), 
that comprises three entities.
The first entity, the \emph{\actor}, is an RL agent that seeks to maximize its accumulated reward in the environment. The second entity, the \emph{\observer}, expects the \actor~to behave in some way and to follow a certain policy, which may differ from the one actually adopted by the \actor. We refer to this  as the \emph{\antpolicy}, and this specifies which actions an observer expects the actor to perform in some set of states.\footnote{Our formalism can be extended to support cases in which the observer anticipates any one of a set of policies to be realized.} The third entity, the \emph{\explainer}, 
has access to a (possibly partial) model of the environment,
to the \antpolicy, and to a set of MDP transforms. The \explainer~seeks a sequence of transforms to apply to the environment such that the \actor's policy in the transformed environment aligns with the \observer's anticipated policy.\footnote{In some settings, the \actor~and \explainer~may represent the same entity. We use this structure to separate the role of an actor from the attempt to explain its behavior.}

\begin{example}\label{example}

\begin{figure}
  \centering
  \includegraphics[width=11cm]{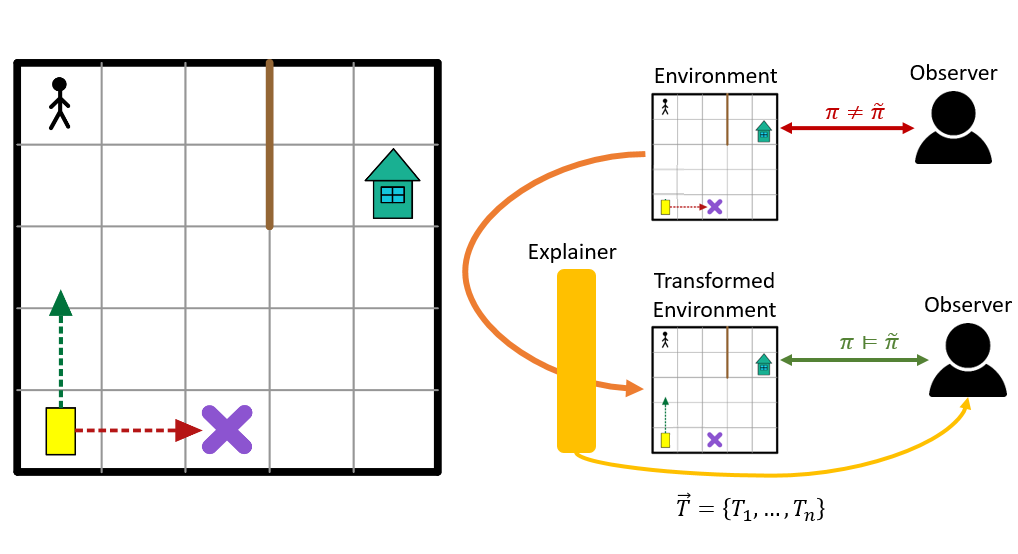}
  \caption{\RLPE, example (left) and model (right). The policy prefix of the taxi (yellow square) is depicted in red and leads to the fuel station (purple mark). The observer, who is not aware of the fuel constraint, anticipates that the taxi heads towards the passenger (the anticipated policy is depicted in green). An explanation is generated by applying a transform that removes the fuel constraint.     
  \label{fig:example_and_model}}
\end{figure}

To demonstrate \RLPEACR, consider Figure \ref{fig:example_and_model}, which depicts a variation of the Taxi domain \cite{dietterich2000hierarchical}. 
In this setting, the \actor~represents a taxi that operates in an environment with a single passenger. %
The taxi can move in each of the four cardinal directions, and pick up and drop off the passenger. The taxi incurs a small cost for each action it performs in the environment, and gains a high positive reward for dropping off the passenger at her destination. There are walls in the environment that the taxi cannot move through. %
The \observer~has a partial view of the environment and knows which actions the taxi can perform and how it can collect rewards. 
With the information available  and the, possibly incorrect, assumptions she makes about the actor's reasoning, %
the \observer~anticipates that the taxi will start its behavior by moving towards the passenger. This description of the anticipated behavior over a subset of the reachable states in the environment is the {\em \antpolicy}. The prefix of this policy is depicted by the green arrow in the figure. 
However, the actual policy adopted by the actor, for which the prefix is represented by the red arrow, is to visit some other location before moving towards the passenger. 
\end{example}

In order to explain the \actor's behavior, the \explainer~applies different action and state space transforms to its model of the environment.
The objective is to find a transformed model in which the \actor~follows the \antpolicy. %\dcp{comment here the MDP is assumed known?} 
We note that our suggested approach can produce meaningful explanations only if the explainer uses transforms that are meaningful to the observer.
In our example, the \explainer~first applies an action transform that allows the taxi to move through walls and trains the \actor~in the transformed environment. Since the policy in the transformed model still does not match the anticipated policy, the \explainer~can infer that the reason for the discrepancy is not the fact that the \observer~may be unaware of the walls in the environment, and therefore this transform would not represent a meaningful explanation. 
As a second attempt, the \explainer~applies a transform that relaxes the constraint that a car needs enough fuel to be able to move, and allows the taxi to move regardless of its fuel level. After training, the \actor's policy in the transformed environment aligns with the \antpolicy.
This indicates the \observer~may not be aware of the fuel constraint, and  does not expect the actor to first drive towards the gas station. This transform is consistent with the discrepancy between the anticipated and actual policies and represents a suitable explanation, as long as this constraint can be conveyed to the observer.  %

Beyond this illustrative example, the ability to understand the ``\emph{anticipation gap}'' (the gap between the anticipated and observed behavior) is important in many applications.
Examples include autonomous driving, where it is critical to know why a vehicle deviates from an anticipated course of action, medical applications, where it is crucial to explain why an AI system recommends one treatment over another, and search and rescue missions, where a robot is moving in an unknown environment with observations that are different from those of its operator and may behave in unpredictable ways. 

The translation of the transform sequence that reconciles the gap between the observer and actor to natural language is beyond the scope of this work. Nevertheless, since the transforms manipulate the underlying MDP model, they incorporate the symbolic information represented by the MDP representation, %\dcp{explain why the transforms you anticipate use `symbolic` information, and what you mean by this} 
and this can reasonably be expected to translate to an intuitive explanation (e.g., notifying the observer about a missing precondition in its model of an action). %
Thus, our approach can be used to automatically generate explanations without compromising generality. Moreover, while we used a single-agent setting to demonstrate the approach, the same ideas can apply to multi-agent settings, where the set of applicable transforms include, in addition to the transforms used for single-agent settings, transforms that deal with the multi-agent aspects of the system (e.g., shared resource constraints).  

The recent interest in explainability in RL has yielded approaches that vary in the kind of questions the explanations are aimed to address and in the methods applied to find them (e.g.,  \cite{pmlr-v97-jiang19a,https://doi.org/10.48550/arxiv.2202.08434,rewDec19,lin2021contrastive,ofra2020local}).
%This variety includes both methods that rely on integrating meaningful representations into the agent's learning process (e.g., \cite{pmlr-v97-jiang19a})
Ours is an example of a post-processing approach, accounting here for settings in which 
 the observer has an anticipated behavior that is not  aligned with the actual behavior, and where the objective is to find an explanation  by transforming the underlying environment to one in which the agent behaves as expected.
 
Typically, post-hoc methods  focus on a particular element of the model and investigate its effect on the agent's behavior.
For example, some propose that 
the reward function  be
 decomposed into an aggregation of meaningful reward types according to which actions are classified
  \cite{rewDec19}, or that human-designed features, such as the estimated distance to the goal, 
  are used to represent action-value functions \cite{lin2021contrastive}. 
In other work, 
human-user studies have been used to extract saliency maps for RL agents in order to evaluate the relevance of features with regard to mental models, trust, and user satisfaction \cite{ofra2020local}, while 
\cite{DBLP:journals/corr/abs-1711-00138,DBLP:journals/corr/abs-1911-11293}
use saliency maps  to produce visual explanations.
Others suggest producing a summary of an agent’s behavior by extracting important trajectories from simulated behaviors  \cite{ofraSumBeh2018}.

Our approach supports arbitrary transforms and abstractions that can be applied to the environment model and combined with any 
learning approach in both single- and multi-agent settings. The variety of transforms that can be used for generating explanations relies on the various methods suggested for expediting planning \cite{geffner2013concise} and RL~\cite{tow06}. Previous work has considered an 
optimal planning agent in a deterministic environment and 
suggested learning a partial model %\dcp{what is a `partial symbolic model`?} 
of the environment and task, and identifying missing preconditions to explain the behavior~\cite{DBLP:journals/corr/abs-2002-01080}. 
We generalize this to stochastic environments with partially-informed RL agents and to arbitrary transforms (beyond only those that consider action preconditions). 

The contributions of this work are threefold. First, we present a novel use of  model transforms and abstractions, formerly mainly used for planning, %
to produce explanations of RL agent behaviors. Second, we present a formal definition of the \RLPE~(\RLPEACR) problem and specify classes of state and action space transforms that can be used to produce explanations. Finally, we empirically demonstrate our approach on a set of standard single-agent and cooperative multi-agent RL benchmarks.

\section{Background}

Reinforcement learning (RL) %
deals with the problem of learning policies for sequential decision making 
in an environment for which the dynamics are not fully known \cite{sutton2018reinforcement}. A common assumption is that the environment can be modelled as a Markov Decision Process (MDP) \cite{bertsekas1995dynamic}, 
typically defined as a tuple $\left<\states, \initState, \actions, \rewardFunc, \transitionFunc, \discountFactor \right>$, where $\states$ is a finite set of states, $\initState \in \states$ is an initial state,  
    $\actions$ is a finite set of actions, 
    $\rewardFunc:\states \times \actions \times \states  \rightarrow \mathbb{R}$ is a Markovian and stationary reward function that specifies the reward $r(s, a, s')$ that an agent gains from transitioning from state $s$ to $s'$ by the
execution of action $a$, 
    $\mathcal{P:S}\times \mathcal{A} \rightarrow \mathbb{P}[\mathcal{S}]$ is a transition function denoting a probability distribution $p(s,a,s')$ over next states $s'$ when action $a$ is executed at state $s$, and
    $\gamma \in [0, 1]$ is a discount factor. %
 In this work we use {\em factored MDPs} \cite{boutilier1999decision}, where each state is described via a set of random variables $X = {X_1, \dots, X_n}$, and where each variable $X_i$ takes on values in some finite domain $Dom(X_i)$. A state is an assignment of a value $X_i \in Dom(X_i)$ for each variable $X_i$. 
To model a multi-agent setting, we use a \emph{Markov game}  \cite{Littman1994}, which generalizes the MDP by including 
{\em joint actions}  $\jointActions =\{\actions^i\}_{i=1}^{n}$ 
representing the collection of action sets $\actions^i$z for each
of the $n$ agents. %\dcpadd{(this is a cooperative setting because the agents share the same reward function)}. 
We will hereon refer to an MDP as the model of the underlying environment, and highlight as needed the specific considerations to a Markov game.  %

A solution to an RL problem is either a {\em stochastic policy}, indicated $\pi:\states \rightarrow \mathbb{P}[\actions]$, representing a mapping from states  $s\in\mathcal{S}$ to a probability of taking an action $a$ at that state, or a {\em deterministic policy}, indicated $\pi:\states \rightarrow \actions$, mapping from states to a single action. The agent's objective is to find a policy that maximizes the expected, total discounted reward.

There are a variety of approaches for solving RL problems \cite{szepesvari2010algorithms,sutton2018reinforcement}, these generally categorized as either {\em policy gradient methods}, which learn a numerical preference for executing each action, {\em value-based methods}, which estimate the values of state-action pairs, or {\em actor-critic} methods, which combine the value and policy optimization approaches. Another important distinction exists between model-based methods, where 
a predictive model is learned, and model-free methods, which learn a  policy directly.
We support this variety by assuming the algorithm that is used by the actor to compute its policy is part of our input. %Our requirement is that each state is characterized by the set of actions that are applicable at that state. \dcp{I don't follow this last sentence. (do you need it?)}%

\section{MDP Transforms}

We use MDP transforms to explain the behaviors of RL agents. Given a large set of possible transforms, an explanation is generated by searching for a set of transforms to apply to the environment's model such that the actor's behavior in the modified model aligns with the observer's expectations. Since the transition from the original to the transformed environment is done by manipulating the symbolic MDP representation of the environment, %\dcp{not clear why you say there is a `symbolic representation of the environ`} 
the difference between the models can help the observer reason about the actor's behavior, thus providing an explanation. 

In this section, we describe various transforms suggested in the literature for expediting planning and RL, and that we apply here for the purpose of explainability. %
We define a {\em transform} as any mapping $\transform: \allMDPs \to \allMDPs$ that can be applied to an MDP to produce another MDP. 
We use the term ``transforms" to refer to various kinds of mappings, including
 ``abstractions" (or ``relaxations") that are typically used to simplify planning, as well as other mappings that may yield more complex environments.  %
Moreover, the set of transforms used for explanation may modify different elements of the MDP instead of focusing on a specific element (e.g, the reward function). We provide some examples of transforms, but our framework is not restricted to particular transforms. %(making it the responsibility of the user to include only meaningful transforms). 
We start by defining transforms that modify the MDP's state space. %\dcp{why `representation of`?}%
\begin{definition}[State Mapping Function]
A \em{state-mapping function}
$\stateAbsFunc: \states \to \states^{\stateAbsFunc}$ maps each
state $s\in \states$, into a state $s^{\prime}\in \states^{\stateAbsFunc}$.
The {\em inverse image} $\stateAbsFunc^{-1}(s^{\prime})$ with $s^{\prime}\in \states^{\stateAbsFunc}$, is the set of states in $\states$ that map to $s^{\prime}$ under mapping function $\stateAbsFunc$.
\end{definition}

When changing the state space of an MDP, we need to account for the induced change to the other elements of the model. For this, we use a state weighting function  %
that distributes the probabilities and rewards of the original MDP among the states in the transformed MDP.

\begin{definition}
[State Weighting Function] \cite{tow06} A {\em state weighting function} of a state mapping function $\stateAbsFunc$ 
is function $\weightingFunc:
\states \to [0, 1]$ where for every $\bar{s} \in \states^\stateAbsFunc$, $\sum_{s \in \stateAbsFunc^{-1}(\bar{s})} \weightingFunc(s)=1$.
\end{definition}

\begin{definition}[State-Space Transform]\cite{tow06}
  Given a state mapping function $\stateAbsFunc$ and a state weighting function $\weightingFunc$, a {\em state space transform} $\transform_{\stateAbsFunc,\weightingFunc}$ maps an MDP $\MDP = \langle \states, \initState, \actions, \rewardFunc, \transitionFunc, \discountFactor \rangle$  to
$\transform(\MDP) = \langle \bar{\states}, \bar{\initState}, \actions, \bar{\rewardFunc}, \bar{\transitionFunc}, \discountFactor \rangle$ where:%\dcp{first two bullets don't seem to use $a$}
\begin{itemize}
    \item $\bar{S}=S^\stateAbsFunc$
    \item $\bar{\initState} = \stateAbsFunc(\initState)$ 
    \item $\forall a\in \actions$,  $\bar{\rewardFunc}(\bar{s},a) =\sum_{s \in \stateAbsFunc^{-1}(\bar{s})} \weightingFunc(s) R(s,a)$
    \item $\forall a\in \actions$, $\bar{\transitionFunc}(\bar{s}, a, \bar{s}')=\sum_{s \in \stateAbsFunc^{-1}(\bar{s})} \sum_{s^{\prime} \in \stateAbsFunc^{-1}(\bar{s}')} \weightingFunc(s) \transitionFunc(s, a, s') $
\end{itemize}
\end{definition}

State-space transforms can, for example, group states together. In factored representations, this can be easily implemented by ignoring a subset of the state features. In Example \ref{example}, a state-space transform can, for example, ignore the fuel level, grouping states that share the same taxi and passenger locations.

Another family of transforms changes the action space. 
\begin{definition}[Action Mapping Function]
 An {\em action mapping function} $\actionAbsFunc: \actions \to \actions^{\actionAbsFunc}$ maps every action in $\actions$ to an action in  $\actions^{\actionAbsFunc}$. The {\em inverse image} $\actionAbsFunc^{-1}(a^{\prime})$ for $a^{\prime}\in \actions^{\actionAbsFunc}$, is the set of actions in $\actions$ that map to $a^{\prime}$ under mapping function $\actionAbsFunc$.
\end{definition}

Various action space transforms have been suggested in the literature for planning with MDPs \cite{mausam2012planning,val20}. Since such transforms inherently bear the MDP's symbolic meaning with regard to the environment and agent, %\dcp{why, and what does this mean?}
a sequence of transforms that yields the anticipated policy can provide a suitable explanation. %\dcp{ok to drop next sentence, i think. explained above} As we note above, the focus here is on automatically finding transforms that bridge this gap, and since they are based on symbolic manipulations of the environment we assume their mapping to natural language is relatively straightforward (but is beyond the scope of this work).

As an example, even if the exact transition probabilities of actions are not fully known, 
%MDP is assumed to be (partially) specified \dcp{why need to say this?}, 
it is possible to apply the {\em single-outcome determinization} transform, where all outcomes of an action are removed (associated with zero probability) except for one,  perhaps the most likely outcome or the most desired outcome \cite{yoon2007ff}. Similarly, the {\em all outcome determinization} transform allows a planner to choose a desired outcome, typically implemented by creating a separate deterministic action for each possible outcome of the original formulation~\cite{yoon2007ff,mausam2012planning}. 
If such transforms yield the anticipated policy, this implies that the observer may  not be aware of the alternative outcomes of an action, or of the stochastic nature of the environment. 
In settings where actions are associated with preconditions, it is possible to apply a  {\em precondition relaxation} transform, where a subset of the preconditions of an action are ignored \cite{DBLP:journals/corr/abs-2002-01080}. For example, for MDPs represented via a factored state space, each action $a$ is associated with a set $\pre (a)$ specifying the required value of a subset of its random variables. A precondition relaxation transform removes the restriction regarding these variables.   
Similarly, it is possible to ignore some of an action's effects, for example by applying a {\em delete relaxation} transform and ignoring an actions' effect on Boolean variables that are set to false \cite{hoffmann2001ff}.
As another example, a {\em precondition addition} transform would add preconditions to an action, perhaps those  that may be considered by the observer by mistake. In all cases, if one or more transforms produce the anticipated policy, a plausible explanation is that the observer is not aware of the preconditions or effects of actions, such as in the setting we describe in regard to fuel in Example~\ref{example}. 

The transforms mentioned above are also applicable to  multi-agent settings. In addition, we can apply multi-agent specific transforms, such as those that allow collisions between agents, or allow for more flexible communication. In a multi-agent extension of our taxi example, an observer may not be aware that taxis cannot occupy the same cell---a discrepancy that can be explained by applying a  transform that ignores the constraint (precondition) that a cell needs to be empty for a taxi to be able to move into it.

\section{Transforms as Explanations}

We formalize the explainability problem as composed of three entities: an {\em \actor}, which is an agent operating in the environment, an {\em \observer}, which is an agent with some anticipation about the behavior of the \actor, and an {\em \explainer}, which is an agent  that  wishes to clarify the discrepancy between the anticipated and actual behaviors.
The input to a {\em \RLPE} (\RLPEACR) problem includes a description of the environment (which may be inaccurate), a description of the behavior (policy) of an RL agent in the environment, the anticipated behavior an observer expects the actor to follow, and a set of possible transforms that can be applied to the environment. 
\begin{definition} [\RLPEACR~Model]
A {\em \RLPE~(\RLPEACR)~model} is defined as $\RLPEModel= \langle \MDP, \agent, \partialPolicy, \transforms \rangle$, where
\begin{itemize}%
        \item $\MDP$ is an MDP representing the environment,
        \item $\agent: \allMDPs \rightarrow \policies$ is the \actor, which is associated with an RL~algorithm that it uses to compute a policy $\policy \in \policies$ , %
        \item $\partialPolicy$ is the
        \antpolicy~the observer expects the actor to follow, and
        \item $\transforms: \allMDPs \rightarrow \allMDPs$ is a finite set of transforms.    
        \end{itemize}
\end{definition}

 We assume the  \actor~is a reward-maximizing \RL~agent\footnote{For the  multi-agent case, instead of a single agent we have a group of agents. All other elements are unchanged.}.
 The anticipated behavior of the \observer~describes what the \observer~expects the \actor~to do in some subset of the reachable states\footnote{The model can be straightforwardly extended to support a set of possible anticipated policies.}.
 Since we do not require the anticipated policy to be defined over all states, we refer to this as a \emph{partial policy}.
The settings of interest here are those in which the actual policy differs from the anticipated policy. %
We denote by $\transformsUniversal$   
the set of all transforms. %\dcp{what does `universal set` mean?}
 Each transform $\transform \in \transformsUniversal$ is associated with a mapping function for each of 
the MDP elements that it alters. We let  $\stateAbsFunc_{\transform}$ and 
$\actionAbsFunc_{\transform}$
denote the state and action mapping functions, respectively (when the MDP element is not altered by the transform, the mapping represents the identity function).
When a sequence of transforms is applied, we refer to the composite state and action mapping that it induces, and define this as follows. 
\begin{sloppypar}
\begin{definition}[Composite State and Action Space Function]
Given a sequence $\transformSeq=\langle \transform_{1}, \dots, \transform_{n} \rangle$, $T_i \in \transformsUniversal$, the {\em composite state space function} of $\transformSeq$, 
is 
$\stateAbsFunc_{\transformSeq}(s) = \stateAbsFunc_{\transform_n}\cdot, \dots,\cdot \stateAbsFunc_{\transform_1}(s)$.
The {\em composite action space function} 
is 
$\actionAbsFunc_{\transformSeq}(s) = \actionAbsFunc_{\transform_n}\cdot, \dots,\cdot \actionAbsFunc_{\transform_1}(s)$.
\end{definition}
\end{sloppypar}

The \explainer~seeks a sequence of transforms that produce an environment where the \actor~follows a policy that corresponds to the \observer 's anticipated policy. Formally, we seek a transformed environment where the actor's policy {\em satisfies} the anticipated policy, i.e., for every state-action pair in the anticipated policy, the corresponding state in the transformed model is mapped to its corresponding action. 
Given a policy $\pi$, we let $\statesFunc(\policy)$ represent the set of states for which the policy is defined.
\begin{definition}[Policy Satisfaction]
Given a partial policy $\pi$ defined over MDP $\MDP = \left<\states, \initState, \actions, \rewardFunc, \transitionFunc, \discountFactor \right>$, a partial policy $\pi^{\prime}$ defined over MDP $\MDP^{\prime} = \left<\states^{\prime}, \initState^{\prime}, \actions^{\prime}, \rewardFunc^{\prime}, \transitionFunc^{\prime}, \discountFactor^{\prime} \right>$, a state mapping function $\stateAbsFunc: \states \to \states^{\prime}$, and an action mapping function $\actionAbsFunc: \actions \to \actions^{\prime} $,  $\pi^{\prime}$ {\em satisfies} $\pi$, denoted $\pi^{\prime} \satisfies \pi$, 
if for every $s\in  \statesFunc(\policy)$, we have $\stateAbsFunc(s)\in  \statesFunc(\pi^{\prime})$ and $ \actionAbsFunc(\pi(s)) = \pi^{\prime}(\stateAbsFunc(s))$. %
\end{definition}

Intuitively, policy $\pi^{\prime}$ satisfies $\pi$ if they agree on the agent's selected action on all states for which $\pi$ is defined. We note that our definition above is suitable only if $\pi(s)$ and $\pi^{\prime}(\stateAbsFunc(s))$ are well-defined, i.e., if the policies are deterministic or, if they are stochastic, a deterministic mapping from states to actions is given (e.g., selecting the maximum probability action). 

%\dcp{add brief interpretation in words}
%
Clearly, for any two policies, there exist state and action mappings that can be applied to cause any policy to satisfy another policy. In order to produce valuable explanations, the input needs to include suitable transforms, i.e., transforms that change the environment in a way that highlights the elements of the model that cause unanticipated behaviors. In addition, and inspired by the notion of a {\em Minimal Sufficient Explanation} \cite{rewDec19}, we want to minimize the change that is applied to the environment. Intuitively, the more similar the original and transformed MDPs are, the better the explanation. We therefore assume the input to an \RLPEACR~problem includes some distance metric, $\MDPdistance: \mathcal{M}\times \mathcal{M} \rightarrow \mathbb{R}_+$, between a pair of MDPs \cite{Song2016}. In our evaluation, the distance  represents the number of {\em atomic changes} that change a single %\dcp{what is this?} to the 
element of the MDP (see the supplementary material for a description of several other distance metrics from the literature).%

The objective of the \explainer~is to find a sequence of transforms that yield an MDP $\MDP^{\prime}$ such that the actor's policy in $\MDP^{\prime}$ satisfies  $\partialPolicy$. Among the sequences that meet this objective, we are interested in sequences that minimize the distance between the original and the transformed MDP. Formally:
\begin{definition}\label{def:RLPE}[\RLPEACR~Problem]
Given a \RLPEACR~model $\RLPEModel$ and a metric function $\MDPdistance : \mathcal{M} \times \mathcal{M}\rightarrow \mathbb{R}_+$ , an \emph{\RLPEACR~problem} seeks a transform sequence $\transformSeq = \langle \transform_1,\dots,\transform_n\rangle$, $\transform_i \in \transforms $, s.t. 
\begin{enumerate}
    \item the \actor's policy $\policy^{\prime}$ in $ \transformSeq(M)$ satisfies $\partialPolicy$, i.e, $\policy^{\prime} \satisfies \partialPolicy$,  and
    \item among the sequences that satisfy (1.), $\transformSeq$ minimizes the distance $\MDPdistance(M, \transformSeq(M))$.
\end{enumerate}

%\dcp{comment that future work could consider finding a transform that gets as close as possible to the anticipated policy?}

\end{definition}

\section{Finding Explanations}

In an \RLPEACR~setting, the \explainer~has access to a set of transforms, but does not know {\em a priori} which transform sequence will produce meaningful explanations. This means that the \explainer~may need to consider a large set of possible transform sequences. This makes a naive approach impractical,
 as the number of transform combinations is exponential in $|T|$. 

\begin{figure}
\RawFloats
  \centering
      \begin{minipage}{0.57\textwidth}
      
  \includegraphics[width=1\textwidth]{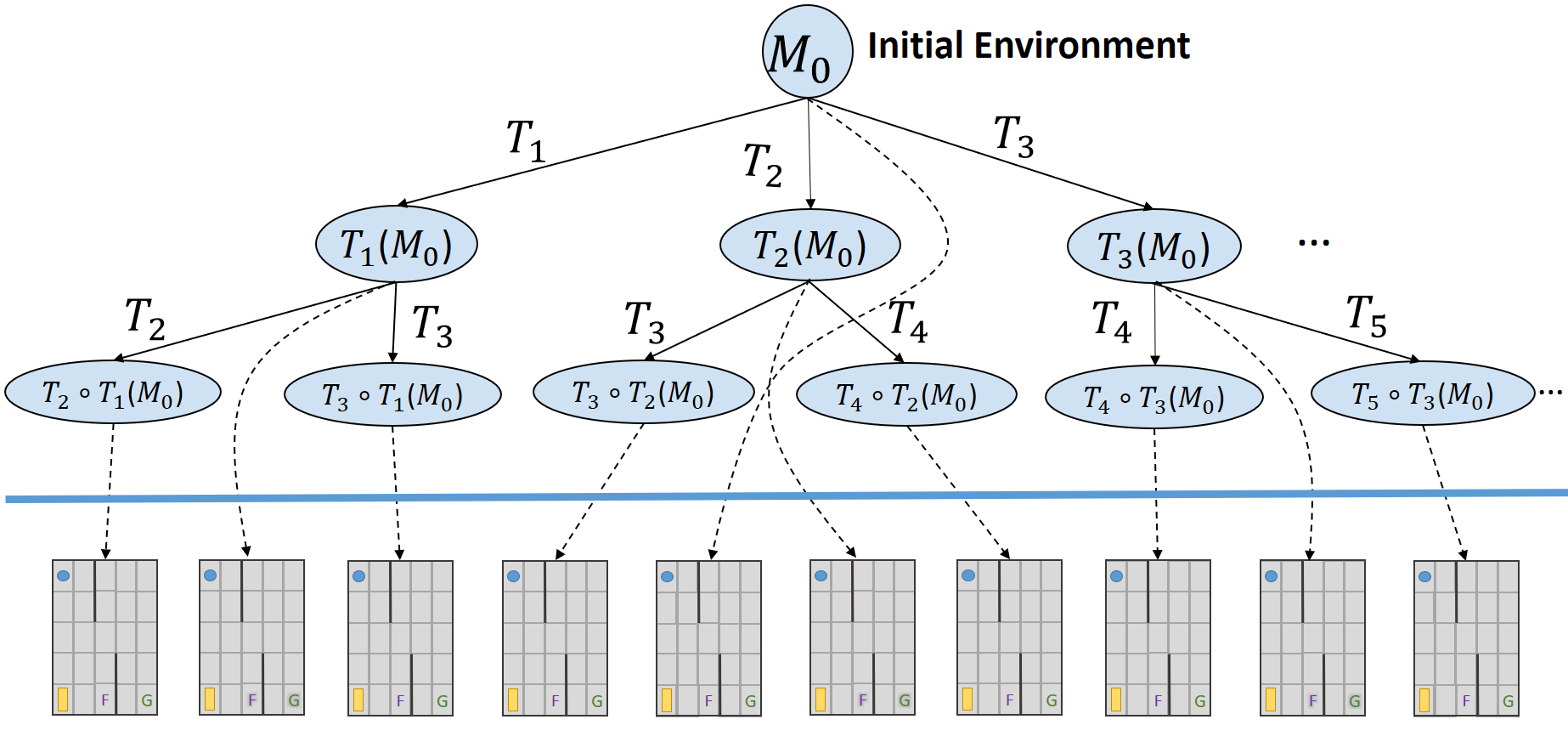}
  \caption{A search for transform-based explanations in the transform space.
      \label{fig:search-transform-tree}}
  \end{minipage}\hfill
  \begin{minipage}{0.4\textwidth}
        \centering
        \includegraphics[width=0.9\textwidth]{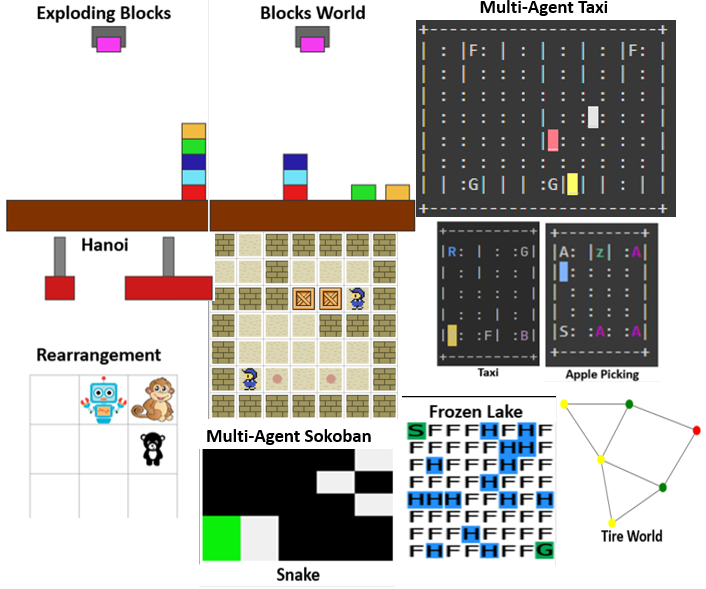}
       \caption{Evaluation domains including single-agent and multi-agent settings.}
       \label{benchmarks}
        \end{minipage}
\end{figure}

%{\bf We chose a symbolic state-space search (A*) because the state space is implicitly given by the set of applicable transforms and a deterministic transition function for which the outcome is known and therefore does not need to be learned. Extending our approach to settings in which these assumptions don't hold, requireling an RL algorithm for the search, is an interesting avenue for future work.}
To address this computational challenge, %
we offer several approaches for expediting the search. Inspired by the search for an optimal MDP redesign in \cite{keren2019}, a basic approach is a Dijkstra-like search through the space of transform sequences. Assuming a successor generator is available to provide the MDP that results from applying each transform, 
the search graph is constructed in the following way. The root node is the original environment. Each edge (and successor node) appends a single transform to the sequence applied to the parent node, where the edge weight represents the distance between the adjacent MDPs according to the distance measure $\MDPdistance$. For each explored node we examine whether the actor's policy in the transformed MDP satisfies the anticipated policy. The search continues until such a model is found, or until there are no more nodes to explore. The result is a transform sequence that represents an explanation. This approach is  depicted  in  Figure \ref{fig:search-transform-tree}, where the top of the figure depicts the search in the transform space and 
 the lower part depicts the MDPs corresponding to each sequence.

The suggested approach is guaranteed to return an optimal (minimum  distance) solution under the assumption that the distance is {\em additive} and {\em monotonic} with respect to the transforms in $T$, in that a transform cannot decrease the distance between the resulting MDP and the original one. From a computational perspective, even though in the worst case this approach covers all the possible sequences, in practice it may find solutions quickly. In addition, in cases where the transforms are {\em independent}, in that their order of application does not affect the result, it is possible to expedite the search by maintaining a closed list that avoids the re-computation of examined permutations. The depth of the search can also be bounded by a predefined fixed number of transforms. 

In spite of these computational improvements, the above solutions require learning from scratch an actor's policy in the transformed environment for each explored node. %\dcp{is this about computing a policy, or learnig a policy?}
One way to avoid this is by preserving the agent's policy in a given environment and using it for bootstrapping re-training in the transformed environment. Another way to expedite the search is to group together a set of transforms and examine whether applying the set leads to a change in the actor's policy. If this compound transform does not change the actor's policy, we avoid computing the values of the individual transforms. This approach is inspired by  pattern database (PDB) search heuristics~\cite{edelkamp2006automated}, as well as the relaxed modification heuristic~\cite{keren2019}. Even though this heuristic approach compromises optimality, it can potentially reduce the computational effort in settings in which aggregation can be done efficiently, such as when transforms have parameterized representations. In our example, if allowing a taxi to move through (all) walls in a given environment does not change the actor's policy, we avoid computing the value of all individual transforms that remove a single wall. Finally, we examine the efficiency of performing a {\em focused policy update}:  when applying a transform, instead of collecting random experiences from the environment and updating the policy for all states, we start by collecting new experiences from states that are directly affected by the transform, and then follow the propagated effect of this change. %\dcp{again, this sounds like planning, not learning} 
In  Example~\ref{fig:example_and_model}, when removing a wall in the taxi domain, we start by collecting experiences and updating the policy of states that are near the wall, and iteratively follow the propagated effect of this change on the policy in adjacent cells.  
\begin{comment}
\begin{figure}
%\vspace{-10pt}
  \centering
  \includegraphics[width=0.7\textwidth]{search_transform_tree.png}
  \caption{A search for transform-based explanations in the transform space.
  \label{fig:search-transform-tree}}
\end{figure}
\end{comment}

\section{Empirical Evaluation}
%{\bf Minor comment 4: The nature of environments (long term reasoning + delayed rewards) chosen is fine. The reason I raised the issue of small environments was to clarify whether the authors have been able to scale these approaches. For instance, I imagine the space of possible transforms may become impractically large when dealing with large state/state-action spaces.}

%{\bf 
%The domains we chose include long-term reasoning and delayed rewards and were therefore challenging for the standard RL we chose for experimentation.
%The only instances we consider in our evaluations are settings in which there is a difference between the anticipated and actual policy. Therefore examining the satisfaction ratio in the original environment is always 0, making it useless to consider the satisfaction ratio relative to the original environment. We will clarify this point.}
%{\bf Results: Also I would prefer detailed explanations like those in Appendix 4 in the main paper.}
%{\bf add a description of how preconditions are learned}
%{\bf Is exhaustive search used for all methods: Base, pre-train, pre-cluster? Or do you prune the transforms after a certain distance is reached?}
%{\bf Q5 Search resources: the search is capped at a fixed and pre-chosen depth.}

The empirical evaluation was dedicated to examining the ability to produce meaningful explanations via MDP transforms and to examining the empirical efficiency of the suggested approaches for finding satisfying explanations. Each \RLPEACR~setting included a description of the underlying environment, the actual policy followed by the actor, and the anticipated policy. We describe each component below, before describing our results\footnote{
Additional results and extensions can be found in the supplementary material. Our complete dataset and code can be found at {\bf https://github.com/sarah-keren/RLPE.git}}.

%\vspace{5pt}
\noindent \textbf{Environments:}
We conducted experiments with $12$ different environments, including both deterministic and stochastic domains and single and multi-agent domains (see Figure \ref{benchmarks}). 
{\em Frozen Lake}~\cite{openaigym} represents a stochastic grid navigation task, with movements in all four cardinal directions and a probability of slipping (and terminating).  
As demonstrated in Example~\ref{fig:example_and_model},
{\em Taxi} is an extension of the similar Open-AI domain (which in turn is based on \cite{dietterich2000hierarchical}), with a fuel constraint that needs to be satisfied in order to move and  actions that correspond to refueling the car at a gas station. {\em Apple-Picking} is our stochastic extension of the Taxi domain: reward is achieved only when picking up a passenger (i.e., an `apple') and  the session can terminate with some probability when an agent encounters a thorny wall.
%In addition, we used six single agent PDDLGym \cite{silver2020pddlgym} domains: Block World, Exploding Blocks, Hanoi, Rearrangement, Snake and Tire World.PDDLGymis a framework that automatically constructs OpenAI Gym environments from PDDL problems descriptions. Observations and actions in PDDLGym are relational: observations are sets of ground relations over objects (e.g. on(plate, table)), and actions are templates ground with objects (e.g. pick(plate)).
%\cite{dietterich2000hierarchical}, which we extend with a fuel constraint.
%We also use Stochastic Apple Picking, in which agents need to collect apples but have a probability of slipping.
We also used seven PDDLGym domains~\cite{silver2020pddlgym}: 
{\em Sokoban, Blocks World, Towers of Hanoi, Snake, Rearrangement, Triangle Tireworld}, and {\em Exploding Blocks}. The PDDLGym framework aligns with the OpenAI Gym interface while allowing the user to provide a model-based relational representation of the environments using PDDL \cite{fox2003pddl2}. This representation is not available to the actor, which operates using standard RL algorithms. For multi-agent domains, we created  a two-agent Sokoban in which agents need to avoid colliding with each other and also provide a \emph{Multi-Taxi} domain that includes multiple taxis that may collide and need to transport multiple passengers\footnote{See \url{https://github.com/sarah-keren/multi_taxi}}.
All these domains have delayed rewards and require multi-step reasoning, making them challenging for standard RL methods.%\dcp{figs 4 and 5 larger if possible}
%
\begin{comment}
\begin{figure}
  \centering
      \includegraphics[width=0.5\textwidth]{domains_images_multi1.png}
      \caption{Evaluation domains including both single-agent and multi-agent settings.
      \label{fig:Domains}}
\end{figure}
\end{comment}

%
\begin{figure}
\RawFloats
  \centering
      \begin{minipage}{.6\textwidth}
      \includegraphics[width=1\textwidth]{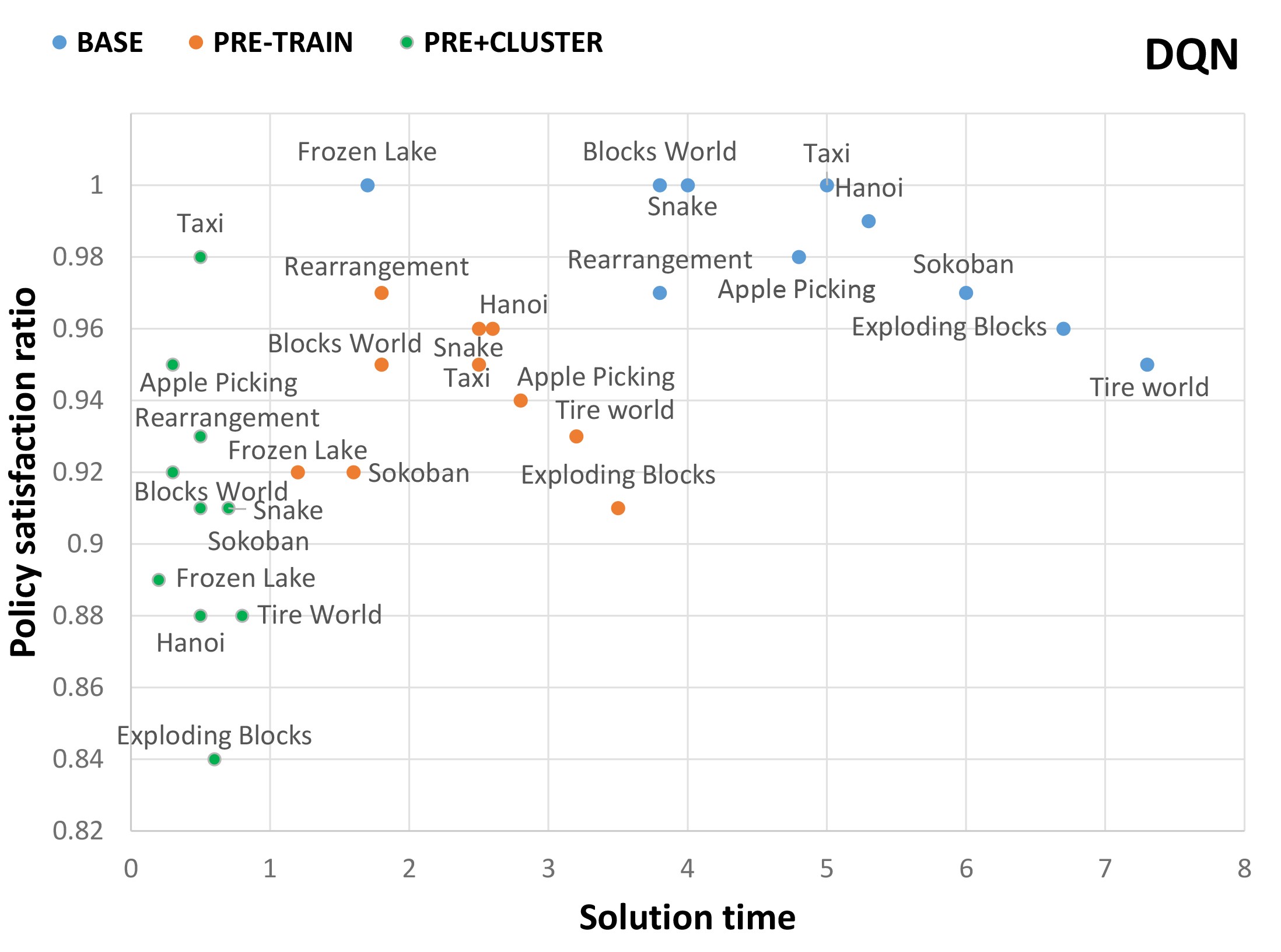}
    \caption{Single-Agent DQN \cite{DBLP:journals/corr/MnihKSGAWR13}: Average running time for finding an explanation for each domain (x-axis), and policy satisfaction ratio, measuring the ratio of instances for which a satisfying policy was found (y-axis).
      \label{fig:resultsdqn}}
  \end{minipage}
  \hfill
   \begin{minipage}{0.6\textwidth}
        \centering
        \includegraphics[width=1\textwidth]{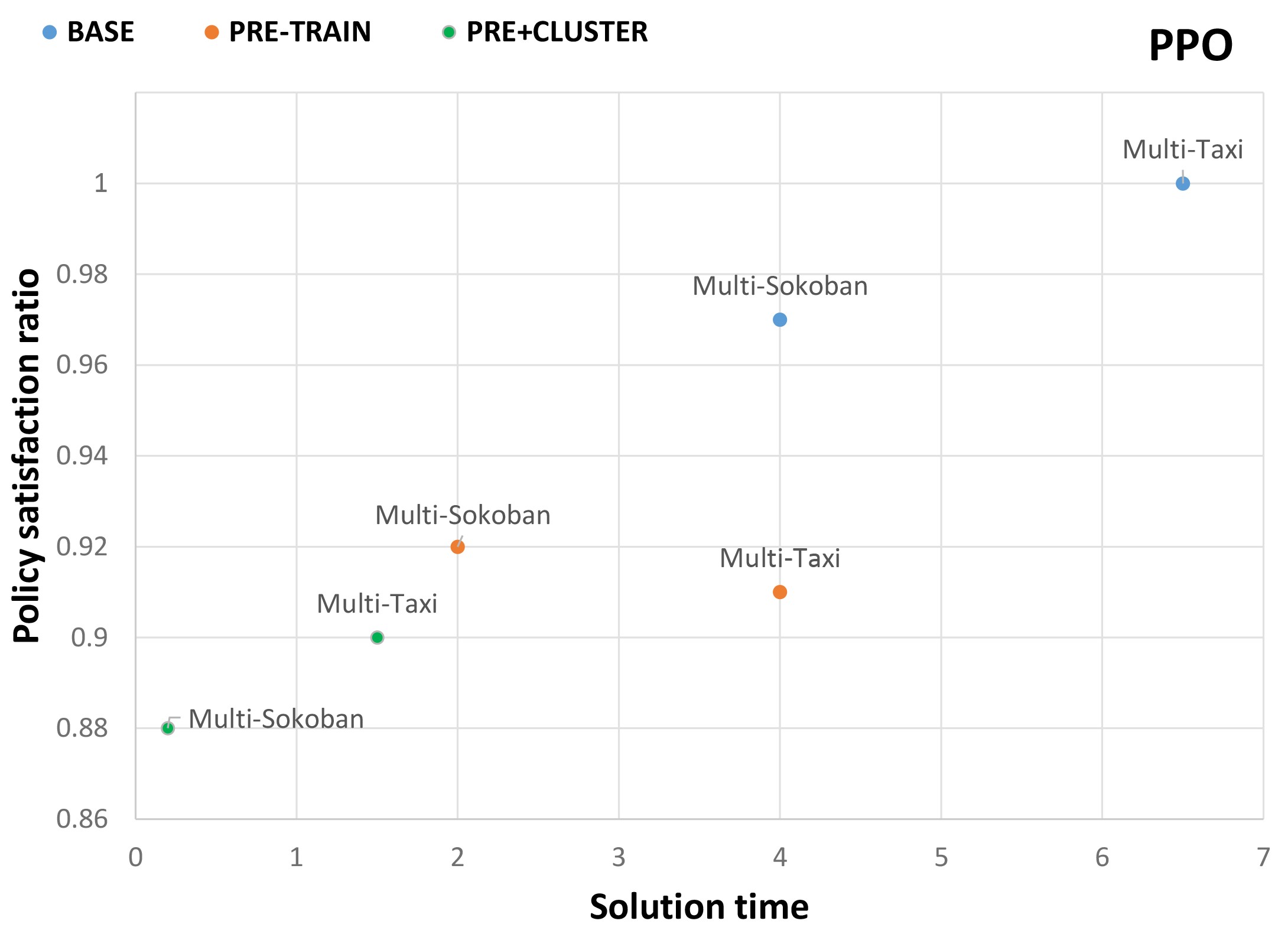}
       \caption{Multi-Agent: Comparing average running time for finding an explanation for the multi-agent domains (x-axis), and policy satisfaction ratio (y-axis). The actors are using PPO \cite{PPO}.
      \label{fig:empirical_summery-multi}}
        \end{minipage}
\end{figure}

\noindent \textbf{Observer:}
We considered a partially informed \observer~that has access to a subset of the environment features. For example, in Taxi 
the observer may be unaware of the fuel constraint or may not be able to see the walls. 
For all environments we assume the observer anticipates that the actor follows a policy that is optimal w.r.t.~the observer's possibly incomplete or inaccurate model. Plans were produced using \cite{mau_magnaguagno_2020_4391071}.%\dcp{why does the observer not need to use an RL algorithm? does A* imply a deterministic environment?}

\noindent
\textbf{Actor:}
For the single-agent settings, we used DQN \cite{DBLP:journals/corr/MnihKSGAWR13}, CEM \cite{szita2006learning}, and SARSA \cite{sutton2018reinforcement} from the keras-rl library\footnote{https://github.com/keras-rl/keras-rl}, as well as Q-learning \cite{Watkins:89}.
For the multi-agent domains, we used PPO \cite{PPO} from keras-rl. 
%We note that our framework can be used with any RL algorithm since it is agnostic to the method used by the actor. 
Agents were trained for 600,000--1,000,000 episodes in each environment, with a maximum of $60$ steps per episode. 

\noindent
\textbf{Explainer:}
We used five  paramterized  transform types: {\em state space reduction} \cite{yoon2007ff}, {\em likely outcome relaxation} \cite{yoon2007ff}, {\em precondition relaxation} \cite{DBLP:journals/corr/abs-2002-01080}, 
{\em all outcome determinization} (for stochastic domains) \cite{Keller_Eyerich_2011}, and \textit{delete relaxation} \cite{hoffmann2001ff}.
Grounding (i.e., the instantiation of the parameterized representations) %dcp{explain} 
was performed automatically for each transform for all environments in which it is applicable.  
Each grounded transform modifies a single action or variable. For the Frozen Lake, Taxi, and Apple Picking domains,  where the dynamics are not defined explicitly, we first learn the transition matrix to generate the \textit{precondition relaxation} transform.
  
We used three methods for searching for explanations. {\em BASE} is a Dijkstra search, {\em PRE-TRAIN} is a Dijkstra search using the learned policy in a given  environment to bootstrap learning in the modified environment, and with a focused policy update to avoid iteratively updating the entire policy. {\em PRE+CLUSTER} extends PRE-TRAIN by computing values of groups of transforms (e.g., applying the delete relaxation to multiple actions) %\dcp{explain} 
and using them to prune individual transforms for which the superset did not change the ratio of states for which the anticipated policy is satisfied. Experiments were run on a cluster using six CPUs, each with four cores and 16GB RAM. We limited the depth of the search tree to three. 

\noindent
\textbf{Results:} %
To assess the ability to produce explanations using environment transforms, 
we measured the {\em \policyMR}~of each transform sequence. This measure is defined as the fraction of states for which the anticipated policy and actor policy agree among all states for which the anticipated policy is defined, i.e., the number of states $s\in  \statesFunc(\policy)$  for which $\stateAbsFunc(s)\in  \statesFunc(\pi^{\prime})$ and $ \actionAbsFunc(\pi(s)) = \pi^{\prime}(\stateAbsFunc(s))$. 
For distance measure $\MDPdistance$, we used the length of the explanation, i.e., the number of atomic transforms (each changing a single element of the MDP) that were applied. %\dcp{what is an `atomic transform`?} 

Figure~\ref{fig:resultsdqn} gives the results achieved by each method for the single-agent domains and with an actor that uses DQN.
Figure~\ref{fig:empirical_summery-multi} gives the results for the multi-agent settings, with 
 PPO used by the agents. 
Each plot represents, for each domain and each method, the average computation time for finding an explanation (x axis) and the average satisfaction ratio (y axis), i.e., the average ratio of the expected policy that was satisfied before the search exhausted the computational resources.
Results for the single agent domains show that while BASE achieves the highest satisfaction ratio (which is to be expected from an optimal algorithm), its computation time is much higher, requiring more than 7x the time of PRE+CLUSTER in Triangle Tireworld. In contrast, PRE+CLUSTER outperforms all other methods in terms of computation time, still with 84\% success in the worst case domain, and with a maximum average variance of 0.03 over the different domains. The results are similar for the multi-agent settings, where the PRE+CLUSTER approach achieved best run time results on both domains while compromising the policy satisfaction rate by up to 10\%. %\dcp{best run time, and good performance?}.      

\section{Conclusion}

We introduced a new framework for explainability in RL based on generating explanations through the use of formal model transforms, which have previously been primarily used for planning. %
The empirical evaluation on a set of single and multi-agent RL benchmarks illustrates
the efficiency of the approach for finding explanations among a large set of transforms. %\dcp{relatively simple domains, though}

Possible extensions include integrating human users or models of human reasoning into the process of generating anticipated policies and in the process of evaluating the quality of the explanations generated by our methods. In addition, while this work uses a restrictive satisfaction relation that requires a full match between the anticipated policy and the actor's behavior in discrete domains, it may be useful to account for continuous domains and to use more flexible evaluation metrics for satisfaction that allow, for example, finding transforms that get as close as possible to the anticipated policy. Finally, our current account of multi-agent settings focuses on fully cooperative settings and it would be interesting to
extend this framework to account for adversarial domains. 

\section{Acknowledgments}

This research has been partly funded by Israel Science Foundation grant \#1340/18 and by the European Research Council (ERC) under the European Union's Horizon 2020 Research and Innovation Programme (grant agreement no. 740282).

%%%%%%%%%%%%%%%%%%%%%%%%%%%%%%%%%%%%%%%%%%%%%%%%%%%%%%%%%%%%

%\bibliographystyle{unsrt}
%\dcp{check refs are up to date, and check capitalization}
\bibliography{RLPE_NEURIPS}

\section*{Checklist} 
\begin{enumerate}
\item For all authors...
\begin{enumerate}
  \item Do the main claims made in the abstract and introduction accurately reflect the paper's contributions and scope?
    \answerYes{}
  \item Did you describe the limitations of your work?
    \answerYes
    %{See page 2 - last paragraph, page 5 - second paragraph, page 7 - last paragraph, page 10 - last paragraph}
  \item Did you discuss any potential negative social impacts of your work?
    \answerNA{}
  \item Have you read the ethics review guidelines and ensured that your paper conforms to them?
    \answerYes{}
\end{enumerate}

\item If you are including theoretical results...
\begin{enumerate}
  \item Did you state the full set of assumptions of all theoretical results?
    \answerNA{}
        \item Did you include complete proofs of all theoretical results?
    \answerNA{}
\end{enumerate}

\item If you ran experiments...
\begin{enumerate}
    \item Did you include the code, data, and instructions needed to reproduce the main experimental results (either in the supplemental material or as a URL)?
    \answerYes%{See supplemental material}
    \item Did you specify all the training details (e.g., data splits, hyperparameters, how they were chosen)?
    \answerYes%{See supplemental material and section 6}
    \item Did you report error bars (e.g., with respect to the random seed after running experiments multiple times)?
    \answerYes%{See supplemental material}
    \item Did you include the total amount of compute and the type of resources used (e.g., type of GPUs, internal cluster, or cloud provider)?
    \answerYes%{See section 6}
\end{enumerate}

\item If you are using existing assets (e.g., code, data, models) or curating/releasing new assets...
\begin{enumerate}
  \item If your work uses existing assets, did you cite the creators?
    \answerYes%{See \cite{openaigym}, \cite{silver2020pddlgym} and page 9 - footnote 6}
  \item Did you mention the license of the assets?
    \answerNA{}
  \item Did you include any new assets either in the supplemental material or as a URL?
    \answerYes%{See section 6, Apple Picking environment}
  \item Did you discuss whether and how consent was obtained from people whose data you're using/curating?
    \answerNA{}
  \item Did you discuss whether the data you are using/curating contains personally identifiable information or offensive content?
    \answerNA{}
\end{enumerate}

\item If you used crowdsourcing or conducted research with human subjects...
\begin{enumerate}
  \item Did you include the full text of instructions given to participants and screenshots, if applicable?
    \answerNA{}
  \item Did you describe any potential participant risks, with links to Institutional Review Board (IRB) approvals, if applicable?
    \answerNA{}
  \item Did you include the estimated hourly wage paid to participants and the total amount spent on participant compensation?
    \answerNA{}
\end{enumerate}

\end{enumerate}

\end{document}